\documentclass{article}

\usepackage[numbers,square]{natbib}
\usepackage[final]{tackling_climate_workshop_style}

\usepackage[utf8]{inputenc} 
\usepackage[T1]{fontenc}    
\usepackage{hyperref}       
\usepackage{url}            
\usepackage{booktabs}       
\usepackage{amsfonts}       
\usepackage{nicefrac}       
\usepackage{microtype}      
\usepackage{xcolor}         
\usepackage[pdftex]{graphicx}
\usepackage{amsmath}
\usepackage{caption}
\usepackage{balance}
\usepackage{color}
\newif\ifworkinprogress
\workinprogresstrue
\ifworkinprogress
  
  \newcommand{\pv}[1]{\textcolor{green}{\textbf{[Puya] #1}}}
\else
  \newcommand{\sm}[1]{}
  \newcommand{\mt}[1]{}
  \newcommand{\pv}[1]{}
\fi

\title{HyperionSolarNet\\ Solar Panel Detection from Aerial Images}

\author{Poonam Parhar, Ryan Sawasaki, Nathan Nusaputra, Felipe Vergara, \\ {\bf Alberto Todeschini, Hossein Vahabi} 
    \\ \texttt{University of California, Berkeley}\\
    \texttt{\{poonamparhar,rsawasaki,nusaputra137,felipe.vergara,}\\
    \texttt{todeschini,puyavahabi\}@berkeley.edu} 
    }
    
\begin{document}

\maketitle

\begin{abstract}
With the effects of global climate change impacting the world, collective efforts are needed to reduce greenhouse gas emissions. The energy sector is the single largest contributor to climate change and many efforts are focused on reducing dependence on carbon-emitting power plants and moving to renewable energy sources, such as solar power. A comprehensive database of the location of solar panels is important to assist analysts and policymakers in defining strategies for further expansion of solar energy. In this paper we focus on creating a world map of solar panels. We identify locations and total surface area of solar panels within a given geographic area. We use deep learning methods for automated detection of solar panel locations and their surface area using aerial imagery. The framework, which consists of a two-branch model using an image classifier in tandem with a semantic segmentation model, is trained on our created dataset of satellite images. Our work provides an efficient and scalable method for detecting solar panels, achieving an accuracy of 0.96 for classification and an IoU score of 0.82 for segmentation performance.
\end{abstract}

\section{Introduction}
As awareness of the impacts and risks of climate change continue to increase, efforts are being made to reduce greenhouse gas emissions. Within the energy industry, strategies are being deployed to lower carbon emissions by reducing fossil fuel energy sources and integrating renewable energy. Although solar panel production continues to increase, the integration of renewable energy is losing momentum and carbon emission reduction goals are falling short due to wavering and unsupportive policy frameworks \cite{IEA}. Many previous works have focused on solar energy forecasting based on weather condition data \cite{PVNet} and solar panel data aggregated from state agencies \cite{TrackingTheSun}. However, simply predicting the solar energy is not sufficient for policy making and determining expansion strategies. In this paper, we design and evaluate an end-to-end framework for creating a world map of solar panels. Our framework identifies locations and total surface area of solar panels within a given geographic area. This information can help solar companies and analysts identify key market areas to target for further expansion and assist policymakers and environmental agencies in making informed decisions to meet renewable energy objectives.

Many recent works have developed methods for solar panel detection \cite{DeepSolar,Bradbury,Zhuang,Wani,DeepConv}. One such study that produced notable advances in this area was DeepSolar, which was trained on over $350,000$ images resulting in a precision and recall for solar panel detection of $90\%$ and a mean relative error of $2.1\%$ for size estimation of solar panels \cite{DeepSolar}. While Deep Solar’s contributions advanced this area of study, their methods require substantial amounts of imagery data and computational power that are only accessible to ventures with the necessary resources. In addition, the vast amounts of images could not be labeled using a supervised, ground-truth approach. Our work takes an alternative approach to previous studies, which results in increased performance while requiring considerably less supervised data and computational resources.

Many studies in this field deploy a two branch model using a classification model to identify the solar panel images, which are then passed to a computationally-heavy segmentation model for mask prediction. Our work, as shown in Figure \ref{fig:architecture}, builds on these foundational studies and introduces a number of enhancements with the following contributions:
\begin{itemize}
	\item {Implementation of an EfficientNet-B7 classifier in combination with a semantic segmentation model based on U-Net architecture utilizing EfficientNet-B7 encoder-backbone to predict the location and size of PV panels.}
	\item {We use a very small dataset consisting of 1,963 satellite images, achieving an average accuracy of 0.96 and an IoU score of 0.82 for our classification and segmentation models, respectively.}
	\item {We demonstrate that our models can be applied in various new locations to generate more labelled data for classification and segmentation training to further improve the models performance. }
	\item {We provide a user-interface that identifies the location and total area of solar panels within a geographical region in quasi-real-time. Additional information on our web application is provided in Appendix D.}
\end{itemize}

\section{Methods}

\subsection{Data}

Given the lack of available datasets, we built our own dataset of satellite images downloaded using Google Maps Static API. To help train robust models, images are selected with diverse features and collected from across the U.S. Batches of images are pulled from Arizona, California, Colorado, Florida, Hawaii, Idaho, Louisiana, Massachusetts, Nevada, New Jersey, New York, Oregon, Texas and Washington. The images are at zoom levels $20$ and $21$, and of sizes $416$x$416$ and $600$x$600$ pixels. In addition, the dataset contains a mix of both residential and commercial buildings. Early trained models revealed a number of false positives with images of objects that resemble solar panels. In subsequent image downloads, we collected \textit{no\_solar} images containing objects that could potentially be misclassified as solar panels, such as skylights, crosswalks, and sides of tall buildings, as shown in Figure \ref{fig:no_solar_images}.

For classification, images are grouped into \textit{solar} and \textit{no\_solar} categories. As shown in Table \ref{Table 1}, both classes split the data 80$\%$ training and 20$\%$ validation. For semantic segmentation, we used LabelBox \cite{labelbox}, a platform to manually annotate solar panels in images and generate segmentation masks as shown in Figure \ref{fig:annotations}. Table \ref{Table 2} shows the total number of labeled images included in the segmentation dataset, which is equal to the number of images in the solar data in the classification dataset. 

In addition to our training and validation set, we also conducted an experiment on the city of Berkeley, CA, and created a test set. The Berkeley test set consists of random sampling of 10$\%$  of the image tiles from the 8 council districts within the entire city. This set consists of 2,243 images with a distribution of 1,922 \textit{no\_solar} and 321 \textit{solar} class images. We use this test set for evaluating our individual models and the complete pipeline.


\begin{table}[htp]
  \begin{minipage}[t]{.5\linewidth}
      \caption{Classification dataset}
      \label{Table 1}
      \centering
      \begin{tabular}{llll}
        \toprule
             & solar  & no\_solar     & Total \\
        \midrule
        Training & 668 & 1295 & 1963     \\
        Validation & 168 & 324 & 492      \\
        Berkeley Test Set & 321 & 1922 & 2243  \\
        \bottomrule
      \end{tabular}
  \end{minipage} %
  \begin{minipage}[t]{.49\linewidth}
      \caption{Semantic segmentation dataset }
      \label{Table 2}
      \centering
      \begin{tabular}{ll}
        \toprule
        & Total \\
        \midrule
        Training & 668      \\
        Validation & 168       \\
        Berkeley Test Set & 321   \\
        \bottomrule
      \end{tabular}
  \end{minipage}
  
\end{table}


\subsection{Models}

\paragraph{Solar panels detection using image classification}

In this work, we employ Transfer Learning and fine-tune an EfficientNet-B7 to classify satellite image tiles into \textit{solar} and \textit{no\_solar}\ classes.  EfficientNet-B7 achieves the state-of-the-art 84.4$\%$  top-1 and 97.1$\%$  top-5 accuracy on ImageNet with 66M parameters and 37B FLOPS. We fine-tune EfficientNet-B7 on a training set consisting of 668 image tiles containing solar panels, and 1295 image tiles without any solar panels. The performance of the fine-tuned model is evaluated against a validation dataset consisting of 168 image tiles containing solar panels, and 324 image tiles without any solar panels.

In order to train EfficientNet-B7 for solar panels image classification, we replace its top layer with a Global Average Pooling layer and a fully connected prediction layer having a single output node. Besides training these two layers, we fine-tune the trainable variables in all of the layers of EfficientNet-B7. The classification model training plots are shown in Figure \ref{fig:classification_model_training_plots}

To increase variation and diversity of input images for model training, we augment the images by applying  random horizontal flip, translation, rotation, contrast and cropping on them. Refer to Figure \ref{fig:augmentation_examples} for example augmented images. We measure the performance of solar panel image classification using accuracy, precision and recall metrics. The performance results on the validation set are shown in Table \ref{Table 3}. The classification model achieves a mean accuracy of 0.98 with a mean precision and recall of 0.98 and 0.95 respectively.


\begin{table}[htp]
    \begin{minipage}[htp]{.5\linewidth}
      \caption{Classification model performance}
      \label{Table 3}
      \centering
      \begin{tabular}{lllll}
        \toprule
        Accuracy  & Precision  & Recall & F1 Score\\
        \midrule
        0.98 & 0.95 & 0.98 & 0.97      \\
        \bottomrule
      \end{tabular}
    \end{minipage} %
    \begin{minipage}[htp]{.49\linewidth}
     \caption{Segmentation model performance}
     \label{Table 4}
     \centering
     \begin{tabular}{lll}
      \toprule
      IoU Score  & F1 Score \\
      \midrule
      0.86 & 0.92      \\
      \bottomrule
     \end{tabular}
    \end{minipage}
\end{table}


\paragraph{Solar panels size estimation using semantic segmentation}

For semantic image segmentation, we use EfficientNet-B7 as the encoder-backbone to train a U-Net model for segmenting solar panels in satellite images. Refer to \ref{fig:unet} for an example of the U-Net architecture trained for segmenting images having solar panels. We annotated 836 images containing solar panels using the LabelBox platform \cite{labelbox}, and produced their corresponding segmentation masks. We resize all images to a size of 512x512 pixels for training and testing. The semantic segmentation model training plots are presented in Figure \ref{fig:seg_training_plots}, and a few examples of the predicted masks are shown in Figure  \ref{fig:examples_seg_predictions}. Refer to Appendix B for the experimental setup and hyperparameters used for model training. The performance results on the validation dataset are shown in Table \ref{Table 4}. Our segmentation model reports an IoU score of 0.86 indicating an average overlap of 86.0$\%$ between the prediction and the ground truth masks, and a mean F1 Score of 0.92 representing the ratio of the overlap and the combined area of the predicted and the label masks.

\paragraph*{Procedure for estimating total surface area and number of solar panels}
The determination of the solar panel surface area is accomplished by programming a function that accepts the output from the segmentation model and returns the total area of solar panels within a given image, as shown in Figure \ref{fig:size_calc}. The segmentation model produces a predicted mask for each image denoting the pixels that are classified as belonging to a solar panel. The predicted mask is then resized to the size of the original image. This pixel matrix is fed into the function along with the latitude and zoom of the image, which are used to derive the size representation of each individual pixel. The length per pixel is calculated using the Mercator Projection, which accounts for the differences in length per pixel depending on the zoom and distance from the equator via: 

\begin{equation}
    \text{meters per pixels} = \frac{156543.03392 \ast \cos{\text{(latitude} \ast \frac{\pi}{180}\text{)}}}{2^\text{zoom}}
\end{equation}

The total area of solar panels is calculated by multiplying the count of ones in the matrix by the area per pixel value. In turn, the number of solar panels is calculated by dividing the total solar panel area by 17.6 ft\textsuperscript{2},\ which is the area of a standard PV panel.

\section{Results}

We evaluated the classification model, segmentation model and the complete pipeline on the test set created consisting of a random sampling of the images from our Berkeley experiment. 

\paragraph{Classification model evaluation}

The HyperionSolarNet classification model reports a mean accuracy of \textbf{0.96} on the Berkeley test set. Table \ref{Table 5} shows the precision, recall and F1 Scores for both the \textit{solar} and \textit{no\_solar} classes. With 1922 no\_solar images in the test set, the model reports a mean precision of 0.98, and a mean recall of 0.97, achieving an average F1 Score of 0.98. In comparison, with 321 solar class images, the classification model achieves a mean precision and recall of 0.82 and 0.91 respectively, and an F1 Score of 0.86. A considerably lower precision for the solar class images indicates that the model makes notable number of false positive predictions.

\begin{table}[h]
 \begin{minipage}[t]{.58\linewidth}
  \caption{Berkeley test set classification}
  \label{Table 5}
  \centering
  \begin{tabular}{lllll}
    \toprule
     Class   & Precision  & Recall  & F1 Score & Support \\
    \midrule
    no\_solar & 0.98 & 0.97 & 0.98 & 1922     \\
    solar & 0.82 & 0.91 & 0.86 & 321      \\
    \bottomrule
  \end{tabular}
  \end{minipage} %
   \begin{minipage}[t]{.4\linewidth}
    \caption{Berkeley test set segmentation}
    \label{Table 6}
    \centering
    \begin{tabular}{ll}
      \toprule
       IoU   & F1 Score  \\
      \midrule
      0.82 & 0.89     \\
      \bottomrule
    \end{tabular}
  \end{minipage}
\end{table}


Figure \ref{fig:test_conf_matrix} in the appendix presents the confusion matrix we obtain by running the segmentation model against the Berkeley test set. To gain a better understanding of where the model is underperforming, we reviewed all the misclassified images. We present a few examples in Figure \ref{fig:examples_misclassified}, where the title of the images indicate their true label, but our model misclassified. There are cases where the model still predicts false positives due to objects such as skylights. Many of the false negatives are due to the model having difficulty detecting PV panels located along the edge of images. As HyperionSolarNet is utilized for various new regions, the misclassified images can be correctly labeled and utilized as training data to improve the model further.

\paragraph{Segmentation model evaluation}

In the Berkeley test set, we had 321 solar-panel images. We manually annotated those images using the LabelBox platform \cite{labelbox}, and created mask labels for them. We evaluated the HyperionSolarNet segmentation model against these test set images and the model reported an IoU score of \textbf{0.82}. The performance results are presented in Table \ref{Table 6}.

In order to understand the model performance better, we examined the predictions that had the IoU score less than 0.4. A few examples are shown in Figure \ref{fig:examples_less_iou}. We note that in these images, the solar panels are not very clear and distinguishable, and it is difficult even for human eyes to identify these solar panels.

\paragraph{Complete pipeline evaluation}
Using our complete model pipeline, we computed the percentage error in the prediction of size and number of solar panels for the Berkeley test set. Our architecture pipeline makes a percentage error of \textbf{0.7$\%$}, as shown in Table \ref{Table 7} in Appendix C. In addition, using all the image tiles of the entire Berkeley city, we estimated that there are around \textbf{61,480} solar panels in Berkeley city. Detailed results are included in Appendix C.

\section{Conclusion}

While this work provides tools to fight climate change, there are potential negative impacts that should be considered. One primary concern is the unintended use of this tool for surveillance and investigating compliance of solar energy mandates, which would disparately impact low-income communities. In addition, while solar panel expansion is beneficial in the fight against climate change, the negative environmental impacts are yet to be realized. Over the next few decades, solar panels will reach the end of their life span resulting in toxic waste that will need proper disposal. 

There is a heightened urgency to solve the climate change crisis by expanding solar energy. Our work provides solutions to this challenge by leveraging deep learning methods and implementing a two-branch model using an EfficientNet-B7 classifier in tandem with a U-Net and EfficientNet-B7 encoder-backbone semantic segmentation model. The use of this two-branch model to predict solar panel locations and surface area results in improved performance while using a relatively smaller training dataset. 
 
\newpage

\begin{ack}
We thank Colorado Reed, Professor Paolo D'Odorico, and UC Berkeley for their support and valuable advice on this work. 
\end{ack}



\newpage

\appendix
\counterwithin{figure}{section}

\section{Appendix}
\subsection{Background}

With the recent advances in deep learning methods, deep learning models are being extensively used in various computer vision tasks such as image classification, object detection, and image segmentation. The fast developments in these tasks have been further accelerated by Transfer Learning. Using Transfer Learning in a classification and segmentation model pipeline, this work presents a supervised deep learning method to estimate the number and surface area of PV panels from high resolution satellite images.

\begin{figure}[htp]
    \centering
	\includegraphics[width=0.99\linewidth]{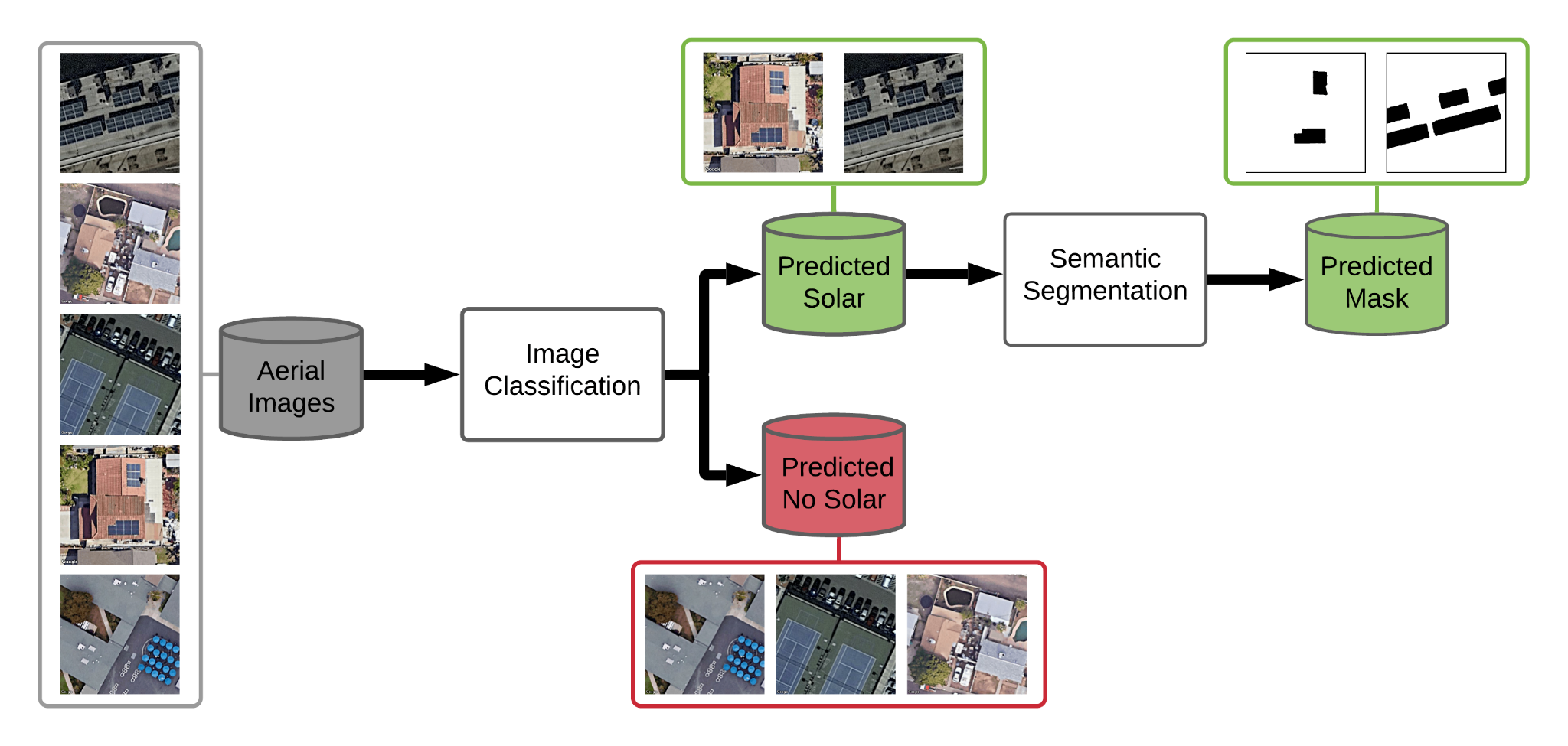}
	\caption{Architecture diagram of the solar panel area prediction pipeline. Satellite images are passed through an image classification model to identify images containing solar panels. The positive predictions from the image classification model are then run through an image segmentation model to identify the pixels belonging to solar panels in an image.}
    \label{fig:architecture}
\end{figure}


\subsection{Image classification}
\begin{figure}[htp]
  \centering
	\includegraphics[width=0.8\linewidth]{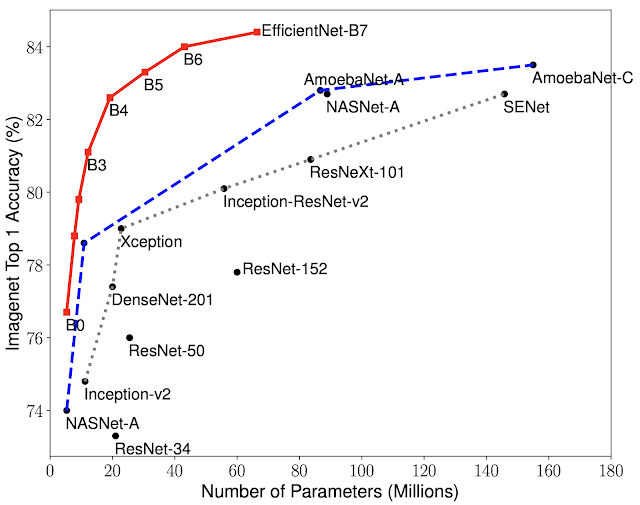}
	\caption{Accuracy scores of various pre-trained image classification models \cite{EfficientNet}.}
	\label{fig:effnet}
\end{figure}

Image classification is the task of classifying images into the groups they belong to. For our case, we define two categories: images containing solar panels and images without any solar panels in them. There are several pre-trained models available for image classification, including but not limited to MobileNet, VGG-16, Inception-v3, RetNet and EfficientNet \cite{EfficientNet}. As shown in Figure \ref{fig:effnet}, there are trade-offs between accuracy and number of model parameters that are to be considered in model selection. We use transfer learning techniques to fine-tune pre-trained classification models for our specific classification task. 


\subsection{Semantic image segmentation}

Image segmentation is a challenging computer vision task where the goal is to classify each pixel of a given image to the specific class or object it belongs to. Semantic segmentation is a specific type of image segmentation where all the pixels representing a particular class are classified as a single entity.

U-Net \cite{unet} is one of the most popular deep learning based semantic segmentation methods. It is a convolutional neural network architecture having Encoder and Decoder layers that down-sample and up-sample the given images to extract features and classify pixels to specific classes. The Encoder or the contraction path is a set of convolutional and max-pooling layers that extract feature maps from the image and reduce its size as it passes through the encoder layers. On the other hand, the Decoder or the expansive path employs transposed convolutional layers along with feature maps from the corresponding encoder layers to restore the resolution and pixel localization information of the image. The layers of the Encoder can be arranged using any of the image classification neural networks, which is called the encoder-backbone of the U-Net architecture. The encoder-backbone also defines how the Decoder layers are built to up-sample the images.

\subsection{Performance metrics}

The performance of our two branch model is evaluated on a validation dataset of 168 \textit{solar} and 324 \textit{no\_solar} images. In addition, we examine model performance on an experiment conducted on the entire city of Berkeley. For the classification model, we use the performance metrics of accuracy, precision, recall, and F1 Score. For the semantic segmentation model, we use the performance metrics of intersection over union (IoU) and F1 Score. The IoU metric is the overlap of the predicted mask and the ground-truth labeled mask divided by their union.


\begin{figure}[htp]
  \centering
	\includegraphics[width=0.8\linewidth]{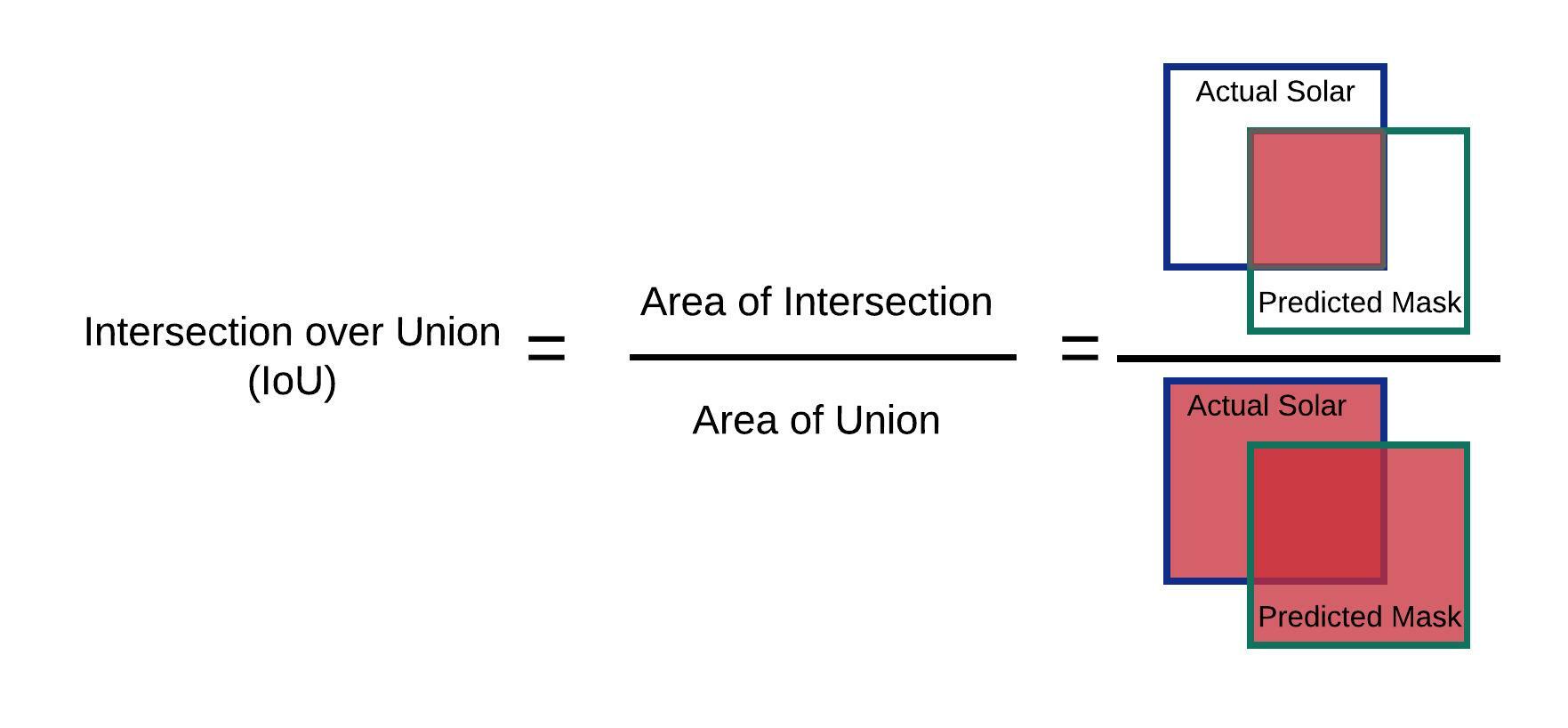}
	\caption{Intersection over Union (IoU) is the area of intersection of the actual object label and the predicted mask divided by their union.}
	\label{fig:iou}
\end{figure}

\pagebreak
\section{Appendix}

\begin{figure}[htp]
  \centering
    \includegraphics[width=0.28\linewidth]{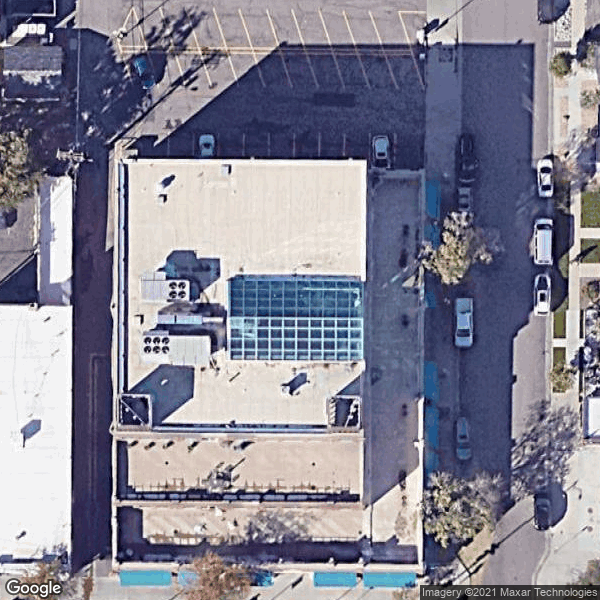}	\includegraphics[width=0.28\linewidth]{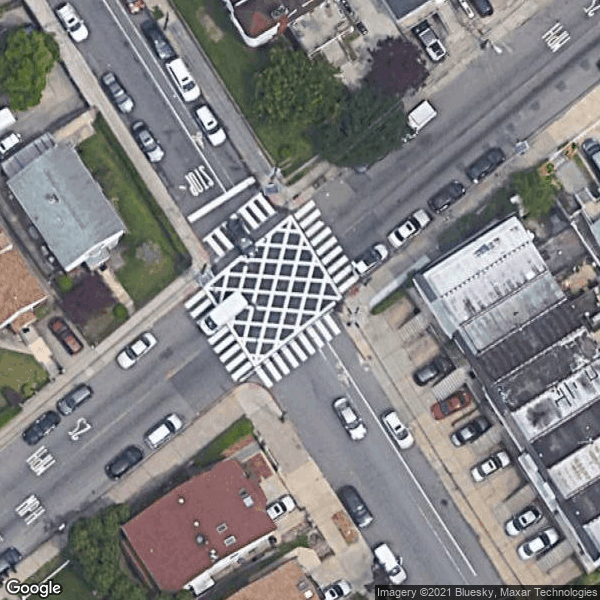}	\includegraphics[width=0.28\linewidth]{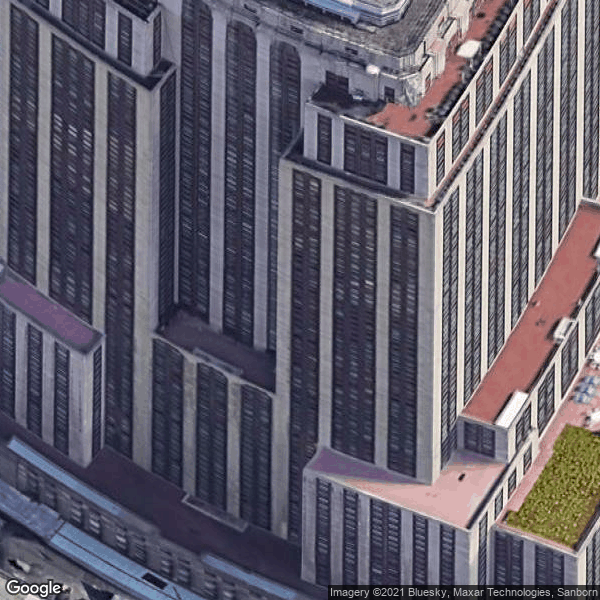}
    \caption{Examples of no\_solar images with objects that resemble solar panels: skylight, crosswalk, side of tall building.}
    \label{fig:no_solar_images}
\end{figure}



\begin{figure}[htp]
  \centering
	\includegraphics[width=0.99\linewidth]{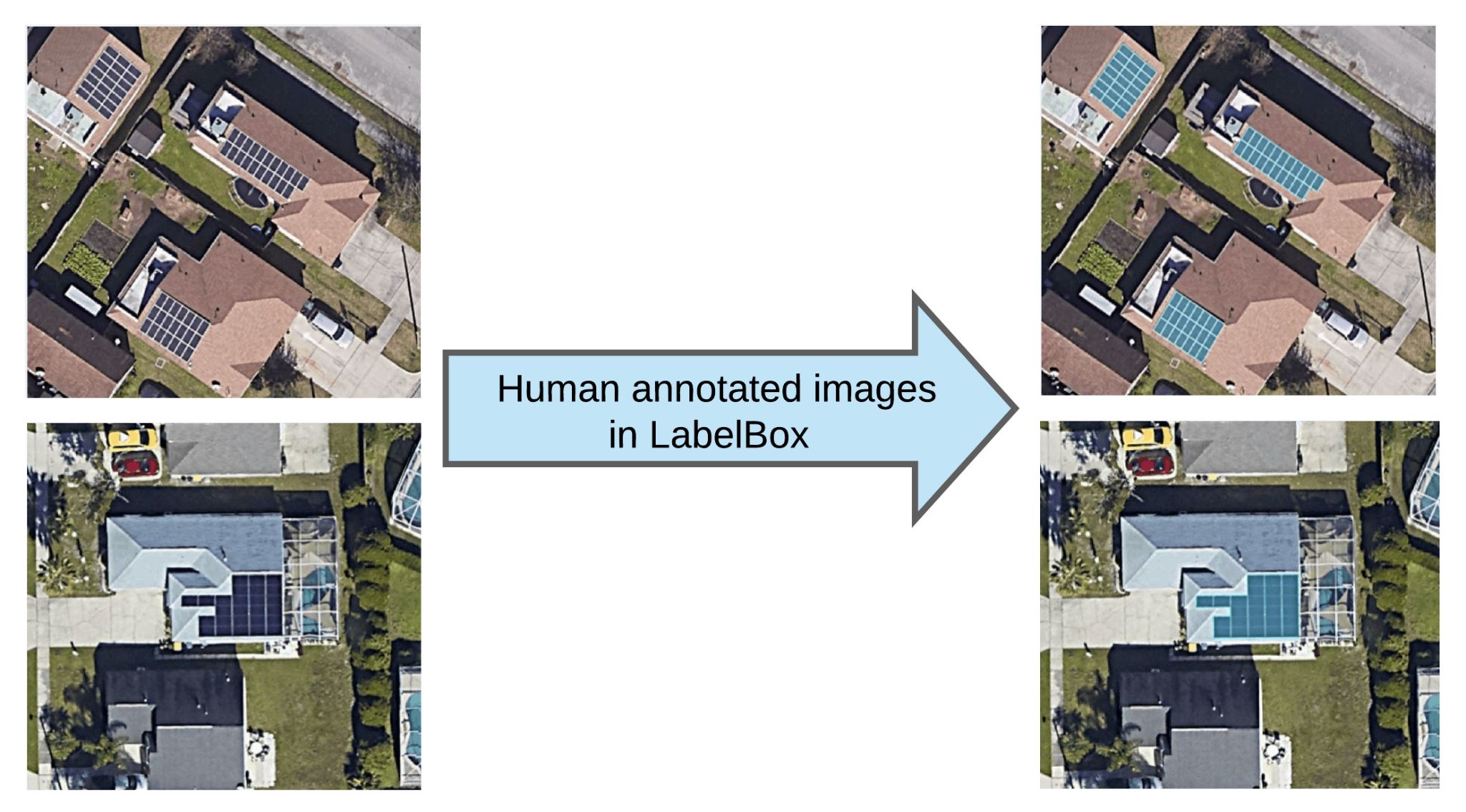}
    \caption{Examples of solar-panel annotations of images in LabelBox.}
    \label{fig:annotations}
\end{figure}



\begin{figure}[htp]
  \centering
	\includegraphics[width=0.99\linewidth]{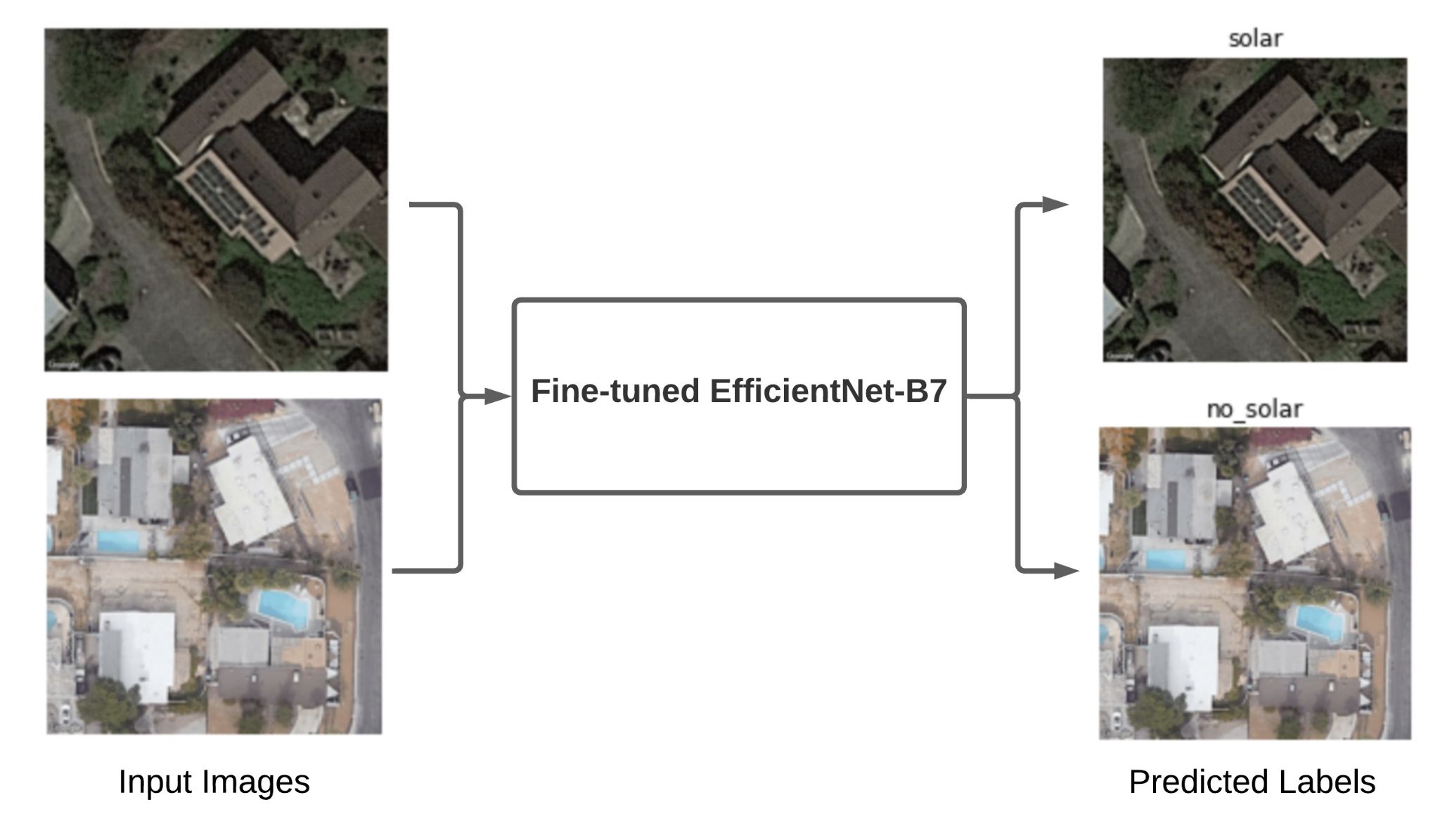}
	\caption{Image classification into solar and no\_solar classes.}
	\label{fig:classification}
\end{figure}



\begin{figure}[htp]
  \centering
	\includegraphics[width=0.9\linewidth]{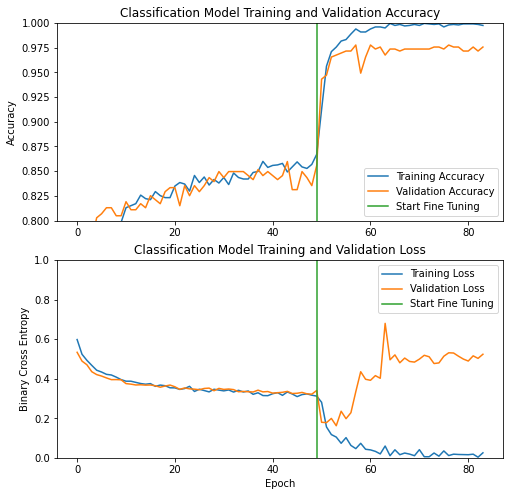}
	\caption{HyperionSolarNet classification model training plots.}
	\label{fig:classification_model_training_plots}
\end{figure}



\begin{figure}[htp]
  \centering	
  \includegraphics[width=0.99\linewidth]{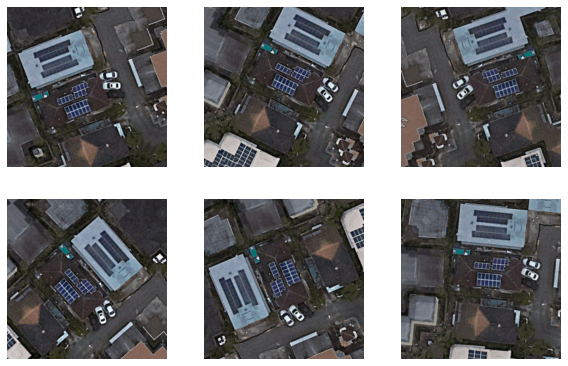}
  \caption{Examples of augmentation on input training images}
  \label{fig:augmentation_examples}
\end{figure}



\begin{figure}[htp]
  \centering
	\includegraphics[width=0.7\linewidth]{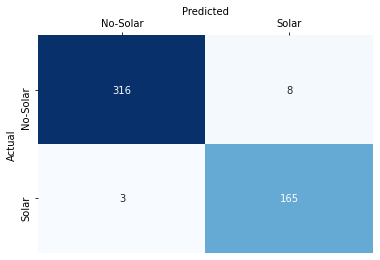}
	\caption{Confusion matrix for the validation dataset with the HyperionSolarNet classification model.}
	\label{fig:val_conf_matrix}
\end{figure}



\begin{figure}[htp]
  \centering
	\includegraphics[width=0.99\linewidth]{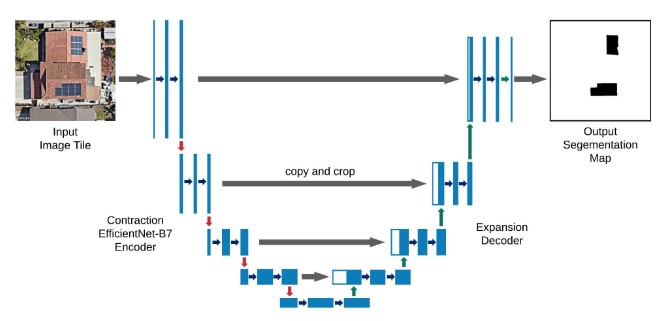}
	\caption{An example of U-Net architecture trained for segmenting images having solar-panels. An image having solar panels is passed through the network, and it produces a segmentation mask for the pixels belonging to solar panels in the image.}
	\label{fig:unet}
\end{figure}



\begin{figure}[htp]
  \centering
    \includegraphics[width=0.99\linewidth]{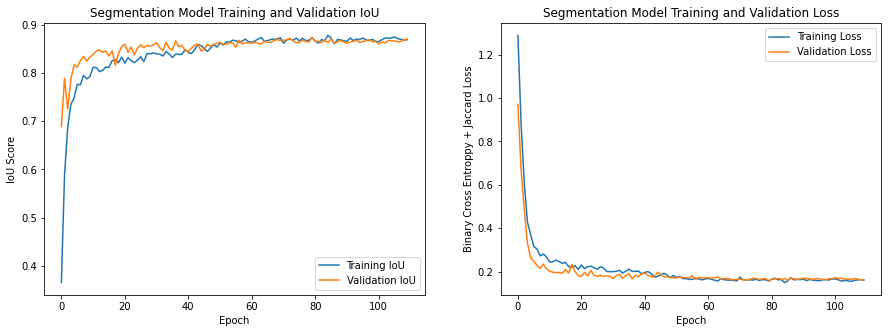}
    \caption{HyperionSolarNet semantic segmentation model training plots}
    \label{fig:seg_training_plots}
\end{figure}


\begin{figure}[htp]
  \centering
    {\textbf{Satellite Image} \hspace{0.2\linewidth} \textbf{Label} \hspace{0.2\linewidth} \textbf{Prediction}}
	\includegraphics[width=0.95\linewidth]{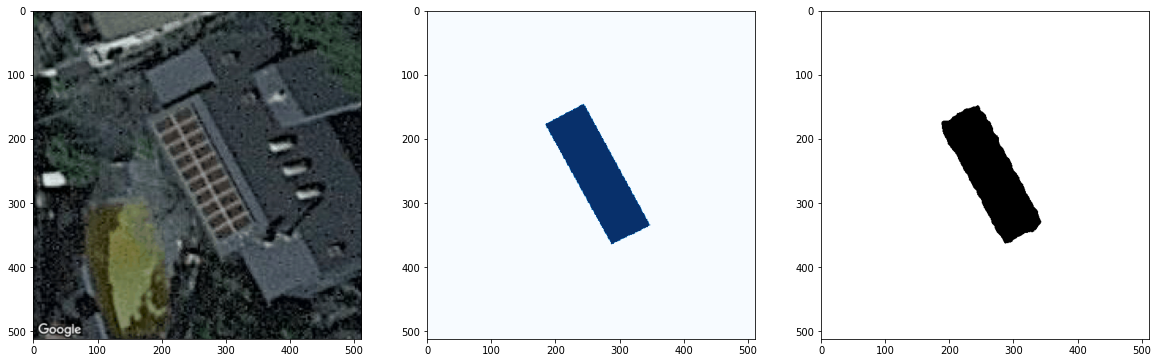}
	\includegraphics[width=0.95\linewidth]{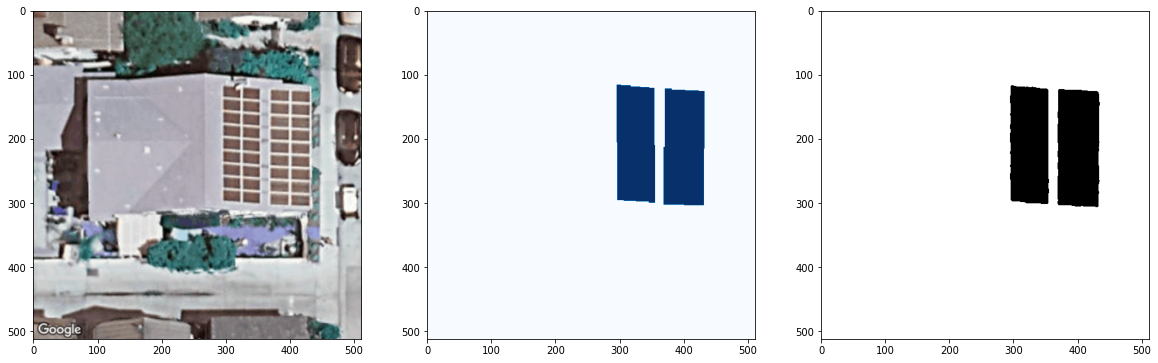}
	\includegraphics[width=0.95\linewidth]{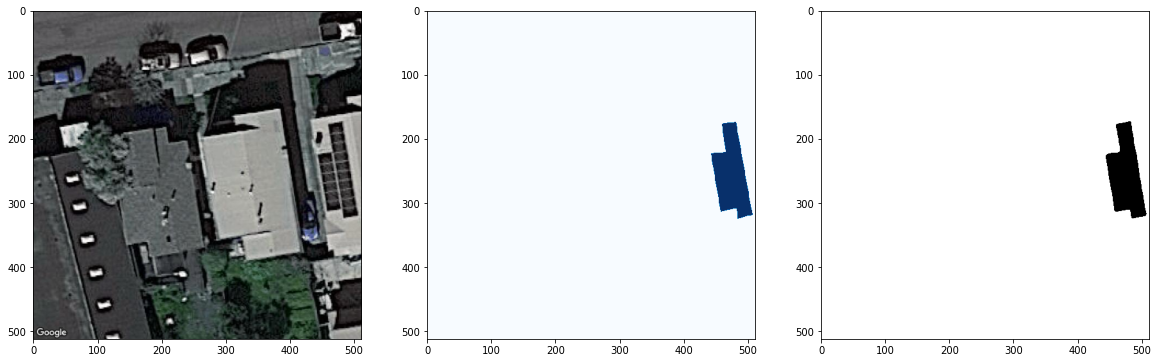}
	\includegraphics[width=0.95\linewidth]{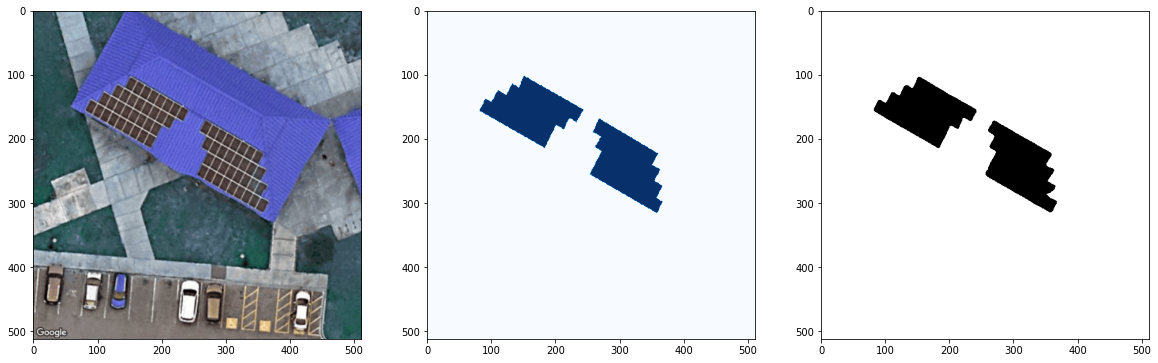}
	\includegraphics[width=0.95\linewidth]{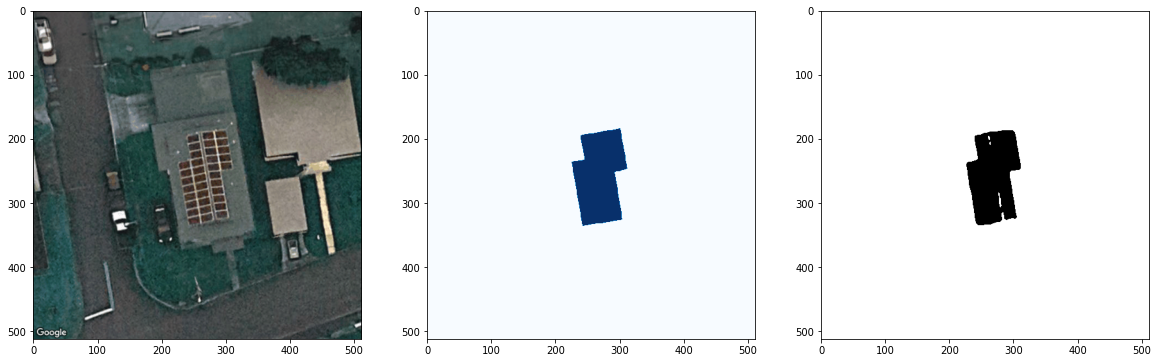}
	\caption{Examples of segmentation mask predictions from HyperionSolarNet, along with the original images and labels.}
	\label{fig:examples_seg_predictions}
\end{figure}

\subsection{Augmentation with Albumentations}
In order to bring variation in our training data images and to increase the dataset size, we augment the images and their segmentation masks. The augmentations include random cropping, flipping, changing brightness and blur, scaling, and grid and optical distortion of images. We use the Albumentations library [\href{https://github.com/albumentations-team/albumentations}{\url{https://github.com/albumentations-team/albumentations}}]\ to perform the augmentations, which enabled us to apply the same augmentation to an image and its corresponding segmentation mask.

\subsection{Segmentation models library}
For training our semantic segmentation model, we use \textit{segmentation\_models} library. It\ offers\ a high level API for training segmentation models, while making 4 different model architectures and 25 backbones available for each architecture. Furthermore, it provides implementation of several segmentation losses (e.g. Jaccard, Dice, Focal) and metrics (e.g. IoU, F1-score), which are very helpful for segmentation model training. The library code is available here: \href{https://github.com/qubvel/segmentation_models}{\url{https://github.com/qubvel/segmentation\_models}}.


\begin{figure}[htp]
  \centering
	\includegraphics[width=0.99\linewidth]{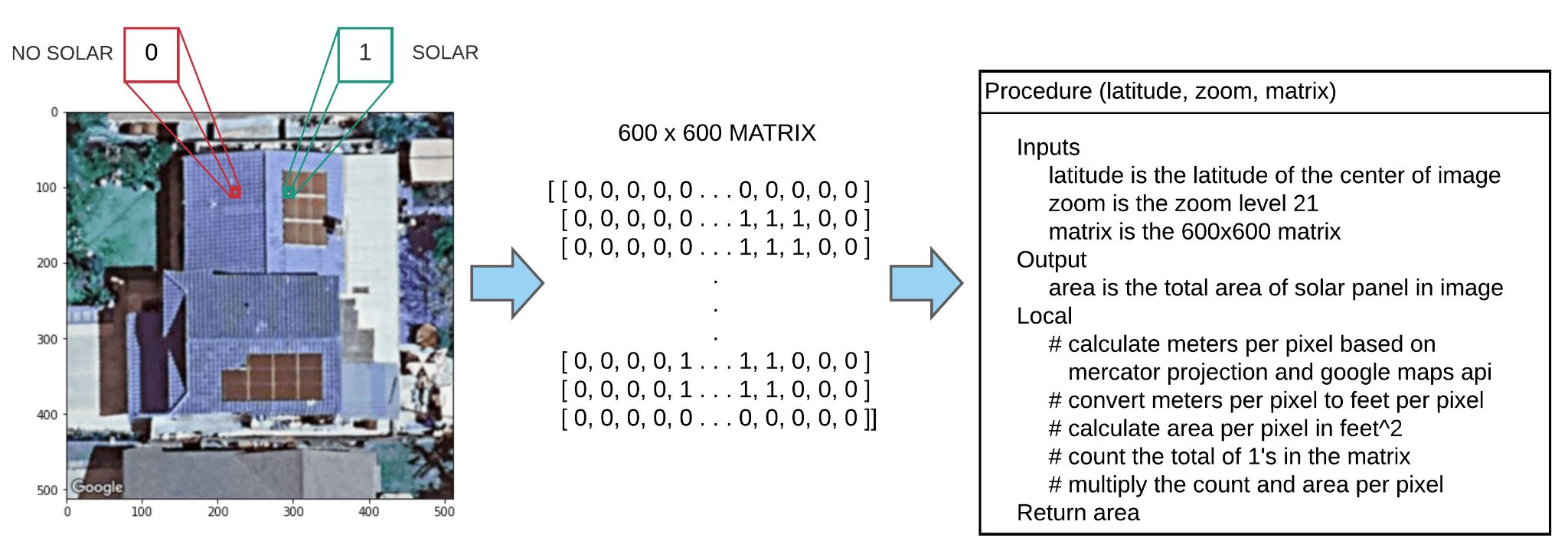}
	\caption{Procedure for calculating the area of solar panel in an image.}
	\label{fig:size_calc}
\end{figure}


\subsection{Experimental setup and hyperparameters}

We trained both of our models on a Google-Colab machine having Tesla V100-SXM2-16GB GPU, and on an AWS g4dn.12xlarge instance having 4 NVIDIA T4 GPUs.

For training and fine-tuning our classification model, we use a starting learning rate of 1e-05 and reduce it by a factor of 0.1 as the validation accuracy reaches a plateau and does not improve for 10 epochs. The base model EfficientNet-b7 has 813 neural network layers. Using a batch size of 8 images, we fine-tune the trainable variables in all of the layers of EfficientNet-b7 for 150 epochs. We use RMSprop as the optimizer and Binary Cross-Entropy as the loss function for our training procedure.

For training the semantic segmentation model, we achieved the best performance using a learning rate of 1e-04 and the Adam optimizer. We reduce the learning rate by a factor of 0.1 as the validation IoU score reaches a plateau. We perform the model training for 150 epochs. For our training loss function, we use the Binary Cross-Entropy and the Jaccard loss. 

\section{Appendix}


\begin{figure}[htp]
  \centering
	\includegraphics[width=0.75\linewidth]{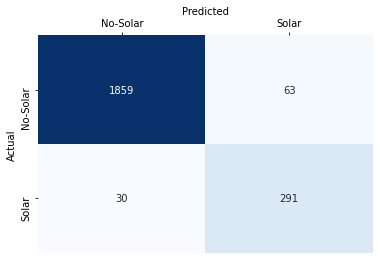}
	\caption{Confusion Matrix for the Berkeley test set with the HyperionSolarNet classification model.}
	\label{fig:test_conf_matrix}
\end{figure}



\begin{figure}[htp]
  \centering
    \includegraphics[width=0.3\linewidth]{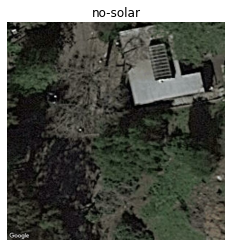}
    \includegraphics[width=0.3\linewidth]{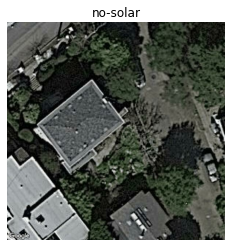}
    \includegraphics[width=0.3\linewidth]{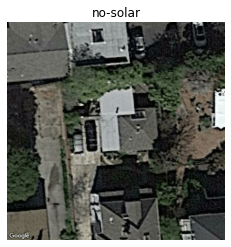}
    \includegraphics[width=0.3\linewidth]{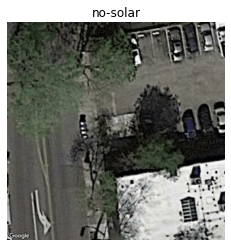}
    \includegraphics[width=0.3\linewidth]{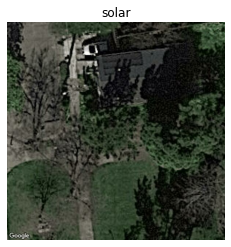}
    \includegraphics[width=0.3\linewidth]{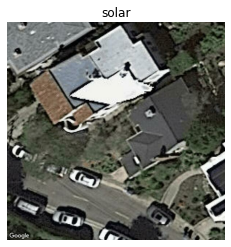}
    \includegraphics[width=0.3\linewidth]{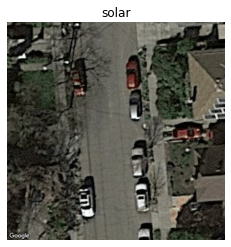}
    \includegraphics[width=0.3\linewidth]{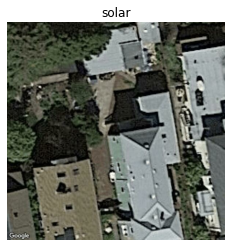}
    \includegraphics[width=0.3\linewidth]{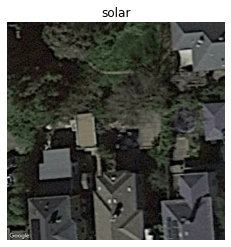}
    \caption{Examples of images misclassified by HyperionSolarNet. The title above the image denotes the actual class, but our model classified as the opposite.}
    \label{fig:examples_misclassified}
\end{figure}


\begin{figure}[htp]
  \centering
    \includegraphics[width=.99\linewidth]{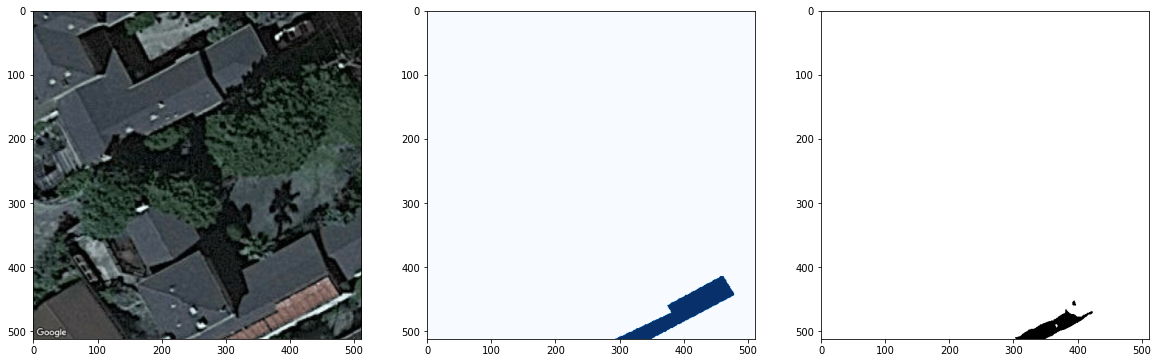}
    \includegraphics[width=.99\linewidth]{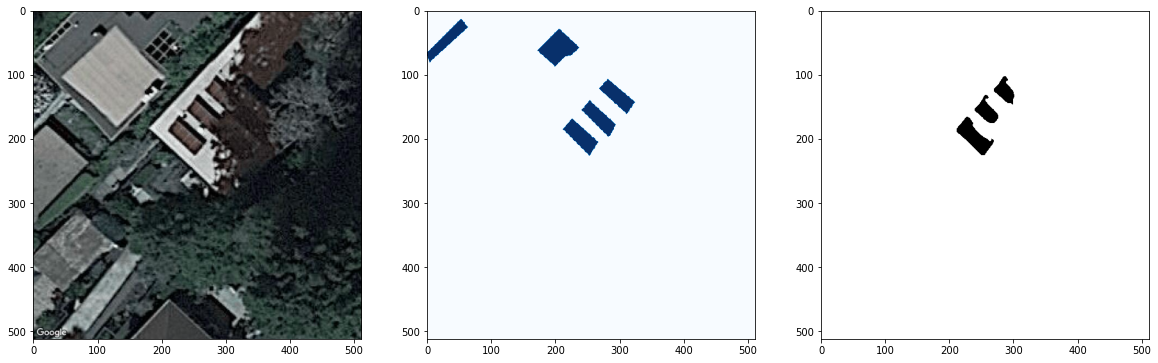}
    \includegraphics[width=.99\linewidth]{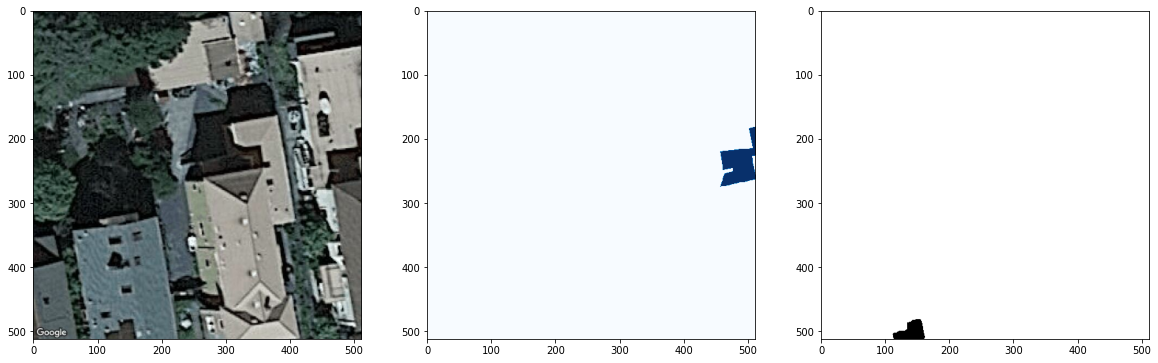}
    \includegraphics[width=.99\linewidth]{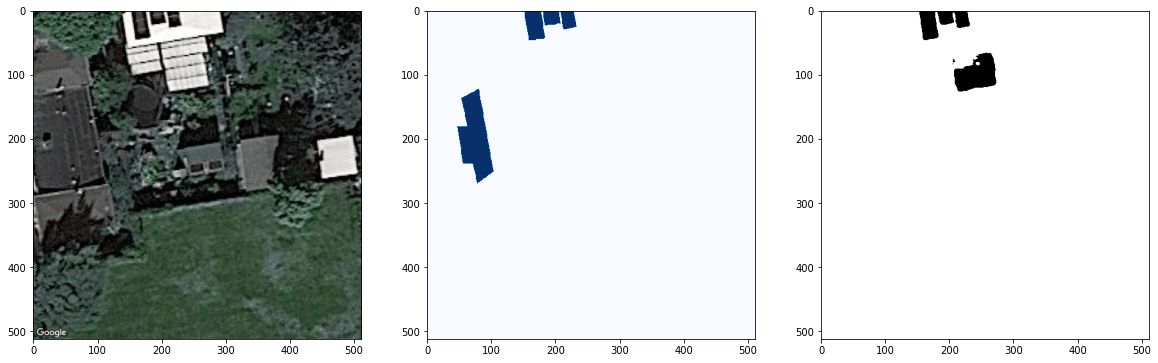}
    \caption{Examples of segmentation predictions with an IoU score < 0.4 from the Berkeley test set}
    \label{fig:examples_less_iou}
\end{figure}


\subsection{Percentage error in the prediction of size and number of solar panels for the Berkeley test set}

Using the label masks of the annotated images in the test set, we calculate the total area and number of solar panels, which we treat as the actual size and number of panels in those images for the purpose of calculating the percentage error in the predictions. Additionally, we estimate/predict the area and number of panels by passing the test set images through our complete classification and segmentation pipeline. Finally, we compare the actual area and the number of panels against the predictions of our pipeline to determine the percent error in prediction for the test set using the following equation. 

\begin{equation}
    \text{Percent Error} = \frac{\text{| Actual - Predicted |}}{\text{Actual}} \cdot 100
\end{equation}

Our architecture pipeline makes an error of 0.7$\%$  in predicting the number and size of solar panels in the Berkeley test set, as shown in Table \ref{Table 7}

\begin{table}[h]
  \caption{Actual and predicted area and number of solar panels for the Berkeley test set}
  \label{Table 7}
  \centering
  \begin{tabular}{lll}
    \toprule
         & Area (sq. ft.)  & Number of Solar Panels \\
    \midrule
    Actual & 101,765.48 & 5,787      \\
    Predicted & 102,609.72 & 5,828      \\
    \bottomrule
  \end{tabular}
\end{table}


\subsection{Estimated solar panels in Berkeley}

We estimated the size and number of solar panels using all the 22,436 image tiles covering the entire Berkeley city. In Table \ref{Table 8}, we share the results.

\begin{table}
  \caption{Estimated total area and number of solar panels in Berkeley}
  \label{Table 8}
  \centering
  \begin{tabular}{ll}
    \toprule
    Area (sq. ft.)  &  Number of Solar Panels \\
    \midrule
    1,082,431.98 & 61,480     \\
    \bottomrule
  \end{tabular}
\end{table}


\section{Appendix}
\subsection{Web application}

We built a web-application that provides users an interactive tool to visualize the output from the classification and segmentation models. The interface allows users to outline a desired geographical area on Google Maps. With the click of a button, HyperionSolarNet splits the region into smaller 600x600 pixel tiles, identifies the latitude and longitude coordinates of each tile containing solar panels, and estimates the total surface area of solar panels. In cases where the desired location is a city boundary or larger, we use a GeoJSON file with the boundary coordinates to preprocess all of the solar panel tiles, locations and surface area within the region.


\begin{figure}[htp]
  \centering
	\includegraphics[width=0.75\linewidth]{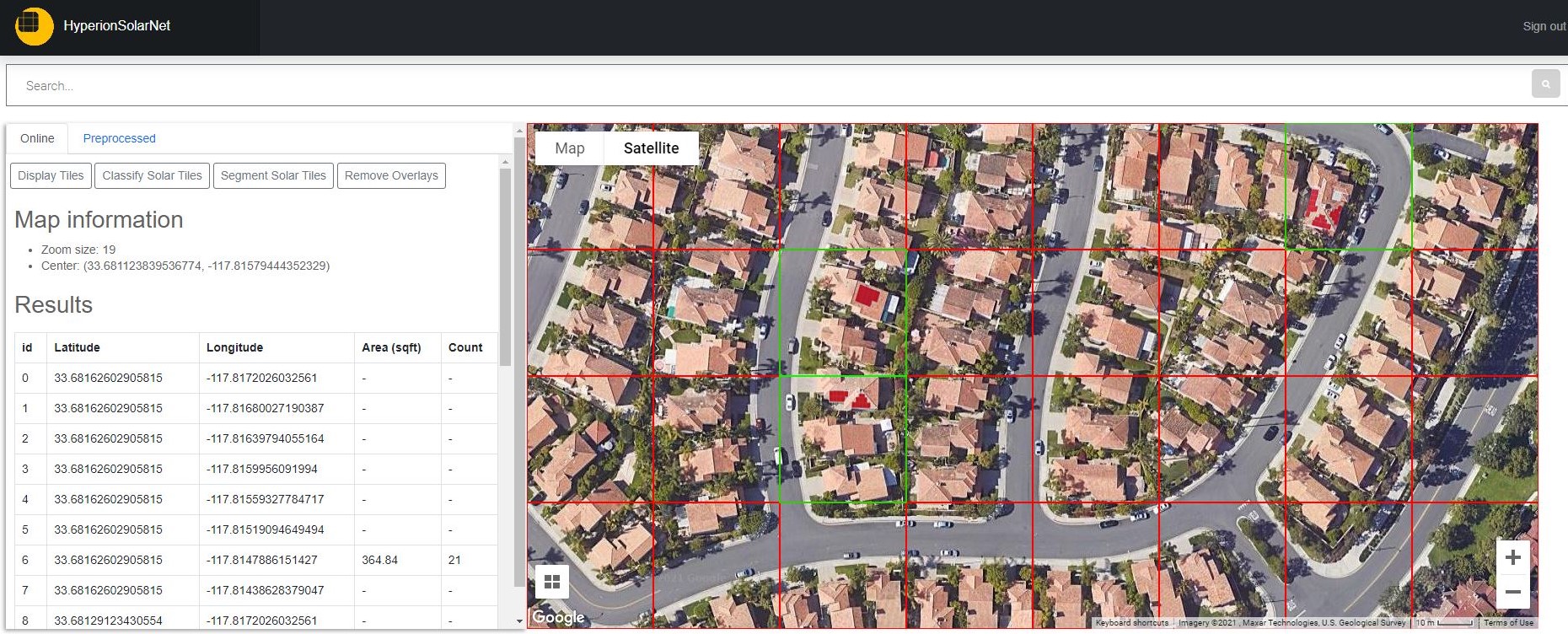}
	\caption{Web application}
	\label{fig:webapp}
\end{figure}


Our application architecture is built on AWS, using EC2 to host our website and S3 to save all static content including models, masked images and GeoJSON files. The website is hosted on a g4dn.xlarge EC2 with T4 GPU and 16 Gb of memory, which is cost effective for inference. The web application is built using Flask, Nginx as a reverse proxy, and Gunicorn as the application server configured to use 3 workers. To achieve inference completely independent from the Flask application, we configure TensorFlow Serving, which provides a server to expose model inference through an API. For our web application on inference, Flask is used as a proxy that connects by calling an API via post to send data to the TensorFlow Serving API which delivers prediction results. As input for inference, we require images for classification and segmentation. The client sends latitude and longitude, height, width and zoom of the map section. Using this information, we use the Google Map API to generate equal size image tiles to use as input for inference, which are converted to numpy arrays. The application returns the classification information for each tile and the segmentation masks overlaid on the image tiles.


\begin{figure}[htp]
  \centering
	\includegraphics[width=0.75\linewidth]{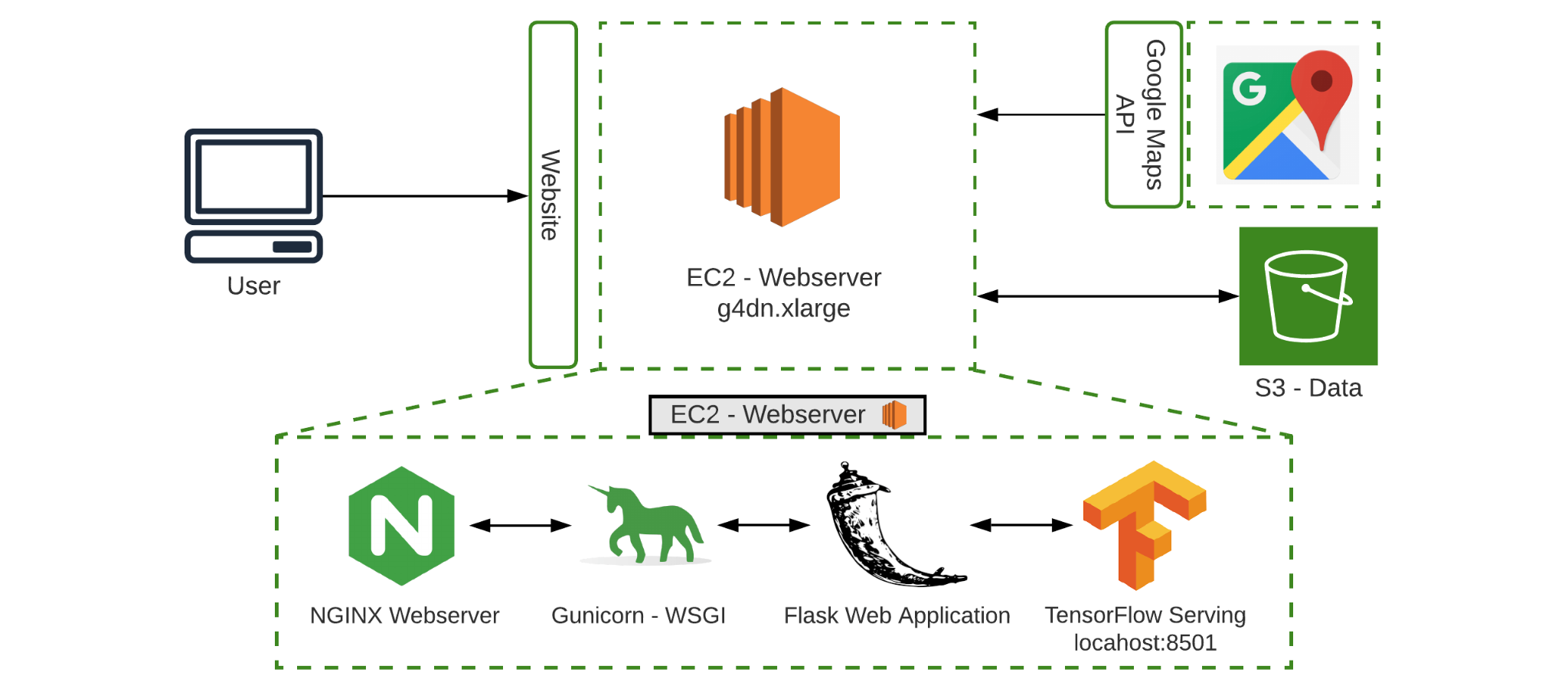}
	\caption{Diagram of HyperionSolarNet system architecture.}
	\label{fig:app_architecture}
\end{figure}


The application has a few limitations mainly due to memory and costs for GPU instances. Due to memory restrictions our minimum zoom capacity for online processing is 19. To reduce the memory usage, we have tested model optimization using TensorRT with the same restrictions. Future enhancements include improving inference performance by testing new optimizations with TensorRT and TensorLight and simplifying the model without sacrificing performance.

\end{document}